\documentclass[conference,letterpaper]{IEEEtran}
\IEEEoverridecommandlockouts

\usepackage{cite}
\usepackage{amsmath,amssymb,amsfonts}
\usepackage{graphicx}
\usepackage{subcaption}
\usepackage{float} 
\usepackage{textcomp}
\usepackage{xcolor}
\usepackage{url}
\usepackage{hyperref}
\usepackage{algorithm}
\usepackage[normalem]{ulem}
\usepackage[noend]{algpseudocode}
\usepackage{cleveref} 
\usepackage[letterpaper, top=0.75in, bottom=1.1in, left=0.625in, right=0.625in]{geometry}
\usepackage{bbm}
\usepackage{balance}

\usepackage{booktabs}
\usepackage{threeparttable}

\def\BibTeX{{\rm B\kern-.05em{\sc i\kern-.025em b}\kern-.08em
    T\kern-.1667em\lower.7ex\hbox{E}\kern-.125emX}}

\begin{document}
\include{math_commands.tex}

\title{A Recommendation System Approach for Interference-Robust Sensor Subset Selection\\
}

\author{
\IEEEauthorblockN{
Kaan Buyukkalayci$^{\dagger}$,
Kyle Pak$^{\dagger}$, Merve Karakas$^{\dagger}$,
and Christina Fragouli$^\dagger$\\
$^\dagger$University of California, Los Angeles\\
Email: \{kaanbkalayci, whilewak, mervekarakas, christina.fragouli\}@ucla.edu
}
}

\maketitle

\begin{abstract}
This paper develops a method for sensor-subset selection for tracking. Prior work showed that low-cost acoustic Received Signal Strength Indicator (RSSI) measurements can be used to recommend subsets of sensor nodes whose expensive sensing modalities, such as cameras, can achieve high tracking accuracy. While efficient, RSSI-based approaches are challenged by acoustic interference. We propose a recommendation-system-inspired framework that instead leverages frequency-band acoustic features and a Two-Tower Multi-Layer Perceptron (MLP) architecture to efficiently score candidate sensor subsets. Experimental results on outdoor vehicle-tracking deployments show that the proposed method can improve accuracy by around 20\%  over the RSSI baseline
while maintaining the low computational overhead required for real-time selective sensing.
\end{abstract}

\section{Introduction}

With the recent growth of edge capabilities in sensing environments, a central problem is the efficient usage of assets with heterogeneous sensing modalities \cite{He2022Collaborative,Siam2025AIoT,Li2023Heterogeneous,marlin2023iobt} as the operating cost of such sensors varies considerably in aspects including, but not limited to, computation, battery, bandwidth, and equipment degradation. For example, the computational and bandwidth cost of processing images can be considerably higher than that of processing audio or vibration data. 
In this paper, we contribute to the problem of {\em group} sensor selection using an approach inspired by recommendation systems. 


Our recent line of work studied the case where a vehicle traverses a field equipped with sensor nodes containing heterogeneous sensing modalities. Using low-cost acoustic sensing, we recommend a small subset of nodes whose higher-cost sensing modalities (e.g., cameras) should be activated \cite{buyukkalayci2025keeping,buyukkalayci2026top,buyukkalayci2025enhancing}.
The recommendation is based on Received Signal Strength Indicator (RSSI) measurements extracted from acoustic data and model-based posterior inference. The appeal of this approach is its simplicity: RSSI measurements are compact and inexpensive to compute, enabling real-time sensor recommendation with minimal communication and processing overhead.

Yet, the use of scalar RSSI measurements also introduces limitations. Because RSSI aggregates all received acoustic energy into a single value, it becomes sensitive to interference from other audio sources in the environment. As a result, interference that alters the acoustic environment can degrade the quality of the recommended subset of nodes. At the same time, acoustic signals contain substantially richer information than what is captured by RSSI alone. Although acoustic sensing cannot replace vision-based sensing, a growing body of work has demonstrated that acoustic information can support meaningful localization and tracking capabilities, particularly in situations where visual sensing is degraded by occlusions, poor lighting, or limited field of view \cite{chakravarthula2023sound}. This suggests that richer acoustic representations may provide more robust recommendations while preserving the advantages of low-cost sensing.

Motivated by these observations, in this paper we seek to extract more information from acoustic measurements without losing the low-complexity, real-time nature that makes selective sensing attractive.
Our goal is still not to replace high-cost modalities such as cameras, but to activate them selectively and only when they are most likely to be useful.

To this end, we revisit  the subset sensor selection problem from a recommendation-systems perspective. 
We note that, at each sensing interval, the network observes a low-cost acoustic description of its current state and must select a small subset of sensing assets whose higher-cost modalities should be activated. This mirrors the structure of modern recommendation systems, where a context is matched against a collection of candidate items and the most relevant items are selected. In our setting, the network-wide acoustic state serves as the context, while candidate sensor subsets serve as the items to be recommended.
This perspective allows us to move beyond explicit target localization and posterior inference, and instead learn a direct mapping from acoustic observations to sensor-subset utility. We propose a recommendation framework based on a Two-Tower Multi-Layer Perceptron (MLP) architecture that learns compact representations of both the network acoustic state and candidate sensor subsets. Similar to retrieval and recommendation systems, the separate representations enable efficient scoring of many candidate subsets while maintaining low online computational cost. 

We evaluate the proposed approach using real-world outdoor vehicle-tracking deployments collected on an IoT sensing platform~\cite{marlin2023iobt}. Experimental results demonstrate that the proposed recommendation framework substantially improves robustness under acoustic interference, up to $20\%$ as compared to the RSSI baseline.
The main contributions of the paper are:
\begin{itemize}
    \item We formulate subset sensor activation as a recommendation problem in which network observations define the context and candidate sensor subsets define the recommendable items.
    \item We introduce a Two-Tower recommendation architecture that learns separate embeddings of network state and sensor subsets, enabling efficient real-time scoring of sensing actions.
    \item We demonstrate that frequency-band acoustic features significantly improve robustness to acoustic interference, increasing accuracy from $80.4\%$ to $98.4\%$ in one of our experiments. 
    \item We show through real-world deployments that the proposed method remains computationally lightweight, requiring sub-millisecond to approximately \(1\,\mathrm{ms}\) of online computation while outperforming prior RSSI-based sensor recommendation approaches.
\end{itemize}

\section{Related Work}
 \textbf{Selective sensing and sensor management for tracking.}
Target tracking in wireless sensor networks is commonly formulated as the problem of estimating a target's location or maintaining its trajectory over time~\cite{li2024online,zhao2002information}. Since continuous sensing, communication, and processing across all nodes can impose substantial battery, bandwidth, and computation costs, prior work has studied resource-aware sensor management mechanisms, including information-driven sensor selection, energy-aware activation, dynamic clustering, and joint allocation of sensing and computation~\cite{zhao2002information,pattem2003coverage,chen2003dynamic}. These methods reduce resource usage while preserving tracking or localization quality. However, these methods primarily optimize tracking or localization accuracy in homogeneous sensing settings.

The selective-sensing problem of keeping a moving target ``on the radar'' was first introduced in~\cite{buyukkalayci2025keeping}, where low-cost sensing modalities are used to recommend a small subset of nodes for activating higher-cost sensing capabilities. This approach was later formalized in~\cite{buyukkalayci2026top}, which provided error bounds for a simplified group subset-selection strategy aimed at identifying the \(p\) most informative nodes. Both works show that estimating compact RSSI measurement models for each node and computing posteriors over possible target locations can provide a computationally efficient and accurate approach to selective sensor activation in sensor networks with acoustic capabilities.  We keep the goal of these works but replace the model-based posterior selection with a learned Two-Tower recommendation model that maps richer multi-band acoustic features directly to subset scores.

\textbf{Two-Tower recommendation models.} Two-Tower (dual-encoder) architectures originate in information retrieval, where separate networks embed a query and a candidate into a shared space for efficient scoring~\cite{huang2013dssm}. They now underpin large-scale recommendation and retrieval systems~\cite{covington2016youtube, yi2019sampling, huang2020embedding}, in which one tower encodes context and the other encodes a candidate, so the candidate embeddings can be precomputed and scored cheaply online. We propose for the first time, to the best of our knowledge, selective sensing in a similar paradigm: a context tower encodes the time-varying acoustic state of the network and a candidate tower encodes a sensor subset, allowing all candidate subsets to be scored within the sensing interval through a learned interaction. This gives us additional flexibility for rarely seen subsets, unlike recent neural sensor-selection methods such as the one introduced in~\cite{strypsteen2025distributed}, which learns input-dependent binary masks end-to-end and does not directly score explicit candidate subsets.

\textbf{Edge AI and collaborative sensing.} Our setting is an instance of edge intelligence for collaborative sensing, in which compute-limited nodes cooperate under tight latency and energy budgets~\cite{kott2016iobt, baek2025, zou2022}. Tiered architectures that use cheap, always-on modalities to trigger expensive sensing have long been used in surveillance sensor networks~\cite{kulkarni2005sens, jurdak2010}; we follow this cheap-triggers-expensive paradigm but learn the triggering policy from data rather than relying on hand-designed rules.


\section{Problem Formulation}
\label{sec:prob_for}

We consider a setting in which a vehicle traverses a field equipped with a set of sensing nodes \(\mathcal V = \{\boldsymbol{s}_i\}_{i=1}^V\) whose positions are known. All nodes in \(\mathcal V\) continuously collect data of a lower-cost modality and transmit them to a central decision-making authority.  We assume that all nodes contain both low- and high-cost modalities; however, the method can be easily implemented in cases where they are not colocated.

The objective is to output a set of nodes \(\mathcal S \subset \mathcal V\) such that \(\mathcal S\) contains the most informative node for high-cost sensing activation. For simplicity, we define this most informative node as the closest node in Euclidean distance to the vehicle, although other metrics can be used by changing the associated utility function depending on the properties of the downstream higher-cost sensing asset.

\textbf{Evaluation metrics.} At each time \(t\), we select a subset \(\mathcal S_t\) using samples from the interval \((t-1,t)\). The objective is to maximize \(P(\boldsymbol{s}_t^* \in \mathcal S_t)\), where \(\boldsymbol{s}_t^*\) denotes the node closest to the target location at time \(t\) while minimizing the subset size \(|\mathcal S_t|\). In our experiments, this is implemented using a fixed subset size. We also report the mean time required for feature construction and real-time inference.

\section{Two-Tower Model}


The model takes two inputs: a context vector describing the acoustic state of the network at a given time, and an action vector specifying a subset of nodes and their relative geometry. These inputs are passed into separate towers, and the resulting embeddings are combined and fed into an MLP head to output a scalar utility score. At inference time, the model scores all candidate subsets and the subset with the highest predicted utility is selected.



\subsection{Feature Construction}

We define the model input as the pair \((\boldsymbol{c}_t,\boldsymbol{a}(\mathcal S))\), consisting of the time-varying context vector \(\boldsymbol{c}_t\in \mathbb{R}^{C}\) and action vector \(\boldsymbol{a}(\mathcal S) \in \mathbb{R}^{A}\). The context vector \(\boldsymbol{c}_t\) describes the acoustic state of the full sensor network at time \(t\) while the action vector \(\boldsymbol{a}(\mathcal S)\) describes the identity and geometry of the candidate subset \(\mathcal S\). Thus, each training example corresponds to a pair \((t,\mathcal S)\) with target value \(u_t(\mathcal S)\).


\vspace{1em}
\noindent\textbf{Context Vector.} We construct the context vector by aggregating node-level coordinate and acoustic band-power features as 
\[
\boldsymbol{c}_t =
\left[
\left\{
\boldsymbol{p}_i,
\boldsymbol{\psi}_{i,t}
\right\}_{i\in\mathcal V}
\right],
\]
where \(\boldsymbol{p}_i\) denotes the normalized coordinates of node \(i\), and \(\boldsymbol{\psi}_{i,t}\) denotes the audio-band power features extracted from the lower-cost acoustic modality at time \(t\).

The audio-band block \(\boldsymbol{\psi}_{i,t}\) contains dB-scale band-power features. Each component is computed as the acoustic power received by node \(i\) within a predefined frequency band over the sampling interval \((t-1,t)\). In our experiments, these bands are listed as:
\vspace{0.8em}

\begin{itemize}
\item[]
\begin{tabular}{@{}lll@{}}
\(\bullet\;20\text{--}80~\mathrm{Hz}\) &
\(\bullet\;80\text{--}160~\mathrm{Hz}\) &
\(\bullet\;160\text{--}400~\mathrm{Hz}\) \\

\(\bullet\;400\text{--}900~\mathrm{Hz}\) &
\(\bullet\;900\text{--}2000~\mathrm{Hz}\) &
\(\bullet\;2000\text{--}3500~\mathrm{Hz}\) \\

\(\bullet\;3500\text{--}6000~\mathrm{Hz}\) &
&
\end{tabular}
\end{itemize}
\vspace{0.5em}

The choice of band edges for our purposes is based on a coarse logarithmic partition of the vehicle-acoustic spectrum, where lower frequencies, in which engine and tire-road energy are concentrated~\cite{pascale2024}, are resolved more finely, while higher frequencies are grouped more coarsely. The required computation at each node is limited to coarse spectral-energy extraction over a small number of frequency bands, after which only a short vector of band-power values must be transmitted to the central decision-making authority. Since the resulting context and action representations are low-dimensional, the decision-making authority can score all candidate subsets within the sensing interval.


\noindent\textbf{Action Vector.} We construct the action vector as
\[
\boldsymbol{a}(\mathcal S)
=
\left[
\boldsymbol{m}(\mathcal S),
\boldsymbol{p}(\mathcal S),
|\mathcal S|
\right].
\]
Here, \(\boldsymbol{m}(\mathcal S)\in\{0,1\}^{|\mathcal V|}\) is the binary membership mask of the candidate subset, with \(m_i(\mathcal S)=1\) if \(\boldsymbol{s}_i\in\mathcal S\) and \(m_i(\mathcal S)=0\) otherwise. The vector \(\boldsymbol{p}(\mathcal S)\) stores the normalized coordinates of the selected nodes using a fixed ordering and padding up to the subset-size budget, and \(|\mathcal S|\) records the subset size. Since \(\boldsymbol{a}(\mathcal S)\) depends only on the subset identity and node geometry, it is independent of \(t\) and can be precomputed for all candidate subsets.

\subsection{Architecture and Training Objective}

We train the model to predict a smooth distance-based utility for each candidate subset. For a candidate subset \(\mathcal S\) at time \(t\), let
\[
d_t^{(1)}(\mathcal S) \leq d_t^{(2)}(\mathcal S) \leq \cdots \leq d_t^{(|\mathcal S|)}(\mathcal S)
\]
denote the sorted distances from the vehicle to the nodes in \(\mathcal S\). The utility of selecting \(\mathcal S\) is defined as
\[
u_t(\mathcal S)
=
\sum_{j=1}^{|\mathcal S|}
\frac{w_j}{1+d_t^{(j)}(\mathcal S)/\rho},
\]
where \(\rho>0\) is a distance scaling parameter and \(w_1 \geq w_2 \geq \cdots \geq 0\) are decreasing weights. This utility gives larger reward to subsets containing nodes close to the vehicle with higher weights assigned to the closest selected nodes, and is chosen to align with our closest-node containment metric. However, the framework is not tied to this particular choice. In other deployments, \(u_t(\mathcal S)\) can be replaced by a task-specific utility that reflects the downstream sensing objective such as visibility, expected detection quality, localization accuracy, or the value of activating a particular high-cost modality.


We use a Two-Tower architecture to predict \(u_t(\mathcal S)\) from the pair \((\boldsymbol{c}_t,\boldsymbol{a}(\mathcal S))\) with one tower representing the time-varying network context and the other representing the candidate sensor subset. The context tower maps \(\boldsymbol{c}_t\in\mathbb{R}^{C}\) to an embedding \(\boldsymbol{z}_c(t)=f_c(\boldsymbol{c}_t)\in\mathbb{R}^{E}\) while the action tower maps \(\boldsymbol{a}(\mathcal S)\in\mathbb{R}^{A}\) to an embedding \(\boldsymbol{z}_a(\mathcal S)=f_a(\boldsymbol{a}(\mathcal S))\in\mathbb{R}^{E}\), where \(E\) is the embedding dimension. Each tower is a small fully connected network with hidden width \(H\).

We combine the two embeddings using their concatenation and elementwise product,
\[
\boldsymbol{r}_t(\mathcal S)
=
\left[
\boldsymbol{z}_c(t),
\boldsymbol{z}_a(\mathcal S),
\boldsymbol{z}_c(t)\odot \boldsymbol{z}_a(\mathcal S)
\right]
\in\mathbb{R}^{3E}.
\]
The elementwise product acts as a learned compatibility feature between the current acoustic state and the candidate subset, so the prediction head can model how different subset geometries align with the observed network-wide acoustic pattern. The combined representation is passed through a prediction head \(g\) to produce the predicted utility \(\hat{u}_t(\mathcal S)=g(\boldsymbol{r}_t(\mathcal S))\). The model is trained by minimizing the mean-squared error between the predicted and true subset utilities. The loss function $\mathcal{L}$ is defined as 
\[
\mathcal L
=
\frac{1}{|\mathcal D|}
\sum_{(t,\mathcal S)\in\mathcal D}
\left(
\hat{u}_t(\mathcal S)-u_t(\mathcal S)
\right)^2,
\]
 where \(\mathcal D\) denotes the training set of timestamp--subset pairs \((t,\mathcal S)\). At inference time, the decision-making authority scores each candidate subset and selects \(\widehat{\mathcal S}_t=\arg\max_{\mathcal S}\hat{u}_t(\mathcal S)\).

\section{Experimental Results}
\subsection{Datasets}

We collect data from outdoor experimental deployments in which a target vehicle traverses a field equipped with spatially distributed sensing nodes. Target vehicle GPS measurements are recorded throughout each run and are used to construct the ground truth for distance-based utilities. Each sensing node is based on an NVIDIA Jetson Orin NX platform and records audio using a reSpeaker 4 Mic Array (UAC1.0) enclosed in a plastic housing. The GPS positions of the sensing nodes are also recorded.

During each experiment, the target vehicle follows a nonuniform outdoor trajectory over an approximately \(25\)-minute run, with typical speed on the order of \(3~\mathrm{m/s}\). The sensing nodes operate within an IoT framework~\cite{marlin2023iobt} and continuously transmit acoustic measurements over an MQTT network. The microphone service records multichannel audio at \(16~\mathrm{kHz}\) using \(16\)-bit samples. We synchronize the acoustic streams with the vehicle GPS measurements so that sensor-subset recommendations can be produced at \(200~\mathrm{ms}\) intervals. After preprocessing, each deployment yields on the order of \(7{,}500\) synchronized timestamps with the chronological training split containing roughly \(15\) minutes of data. We evaluate the proposed method on two distinct outdoor deployments and report the results for each. For training, each timestamp is paired with every candidate subset of the fixed recommendation budget, producing timestamp--subset examples separately for each deployment.

\vspace{1em}

\noindent\textbf{Interference-Rich Deployment.}
This deployment uses six sensing nodes and a cargo-van target vehicle in an outdoor area of approximately \(4{,}000~\mathrm{m}^2\). The site contains hilly terrain, buildings, and a nearby construction zone. During data collection, additional acoustic interference was present intermittently, including human speech, wind, and occasional audio from nearby passing vehicles. These interference sources are sporadic rather than continuously active, allowing us to evaluate robustness under realistic nonstationary acoustic conditions.

\begin{figure}
    \centering
    \includegraphics[width=\linewidth]{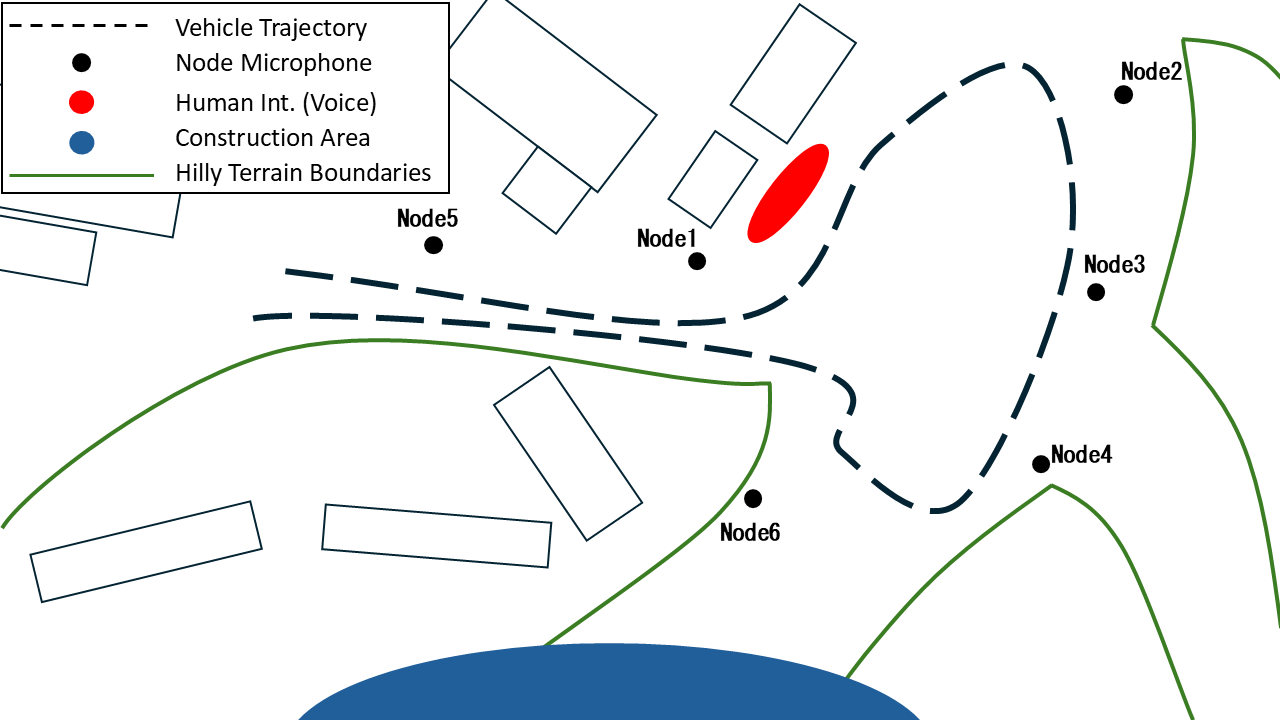}
    \caption{Sketch of the interference-rich experiment site.}
    \label{fig:wp_map}
    \vspace{-1em}
\end{figure}

The target vehicle follows a nonuniform trajectory through the site, rather than moving along a single straight path. The vehicle path, sensing-node locations, buildings, terrain boundaries, and representative interference sources are shown in Fig.~\ref{fig:wp_map}.

\begin{figure}
    \centering
    \includegraphics[width=0.9\linewidth]{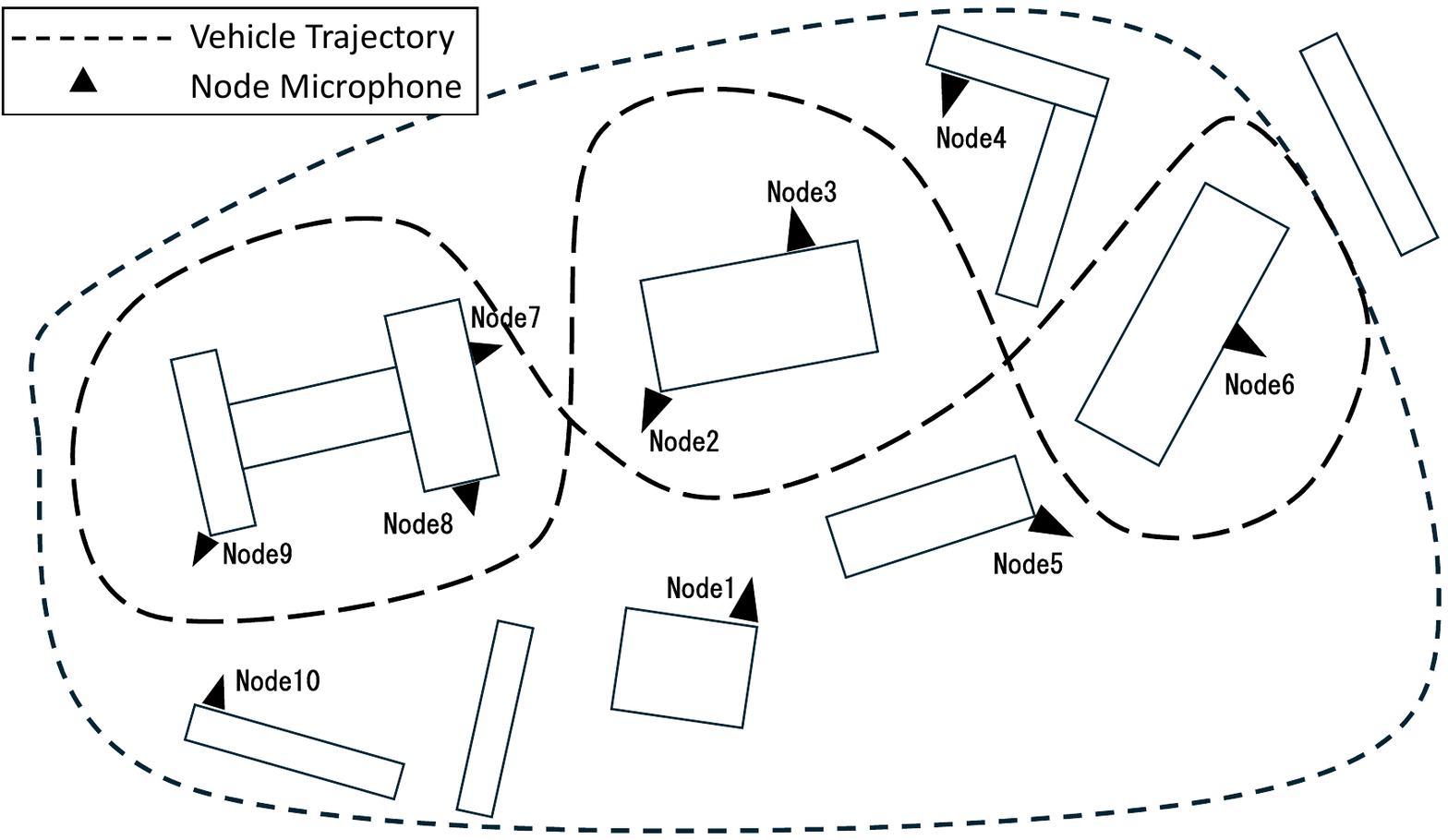}
    \caption{Sketch of the open-field experiment site.}
    \label{fig:gq_map}
    \vspace{-1em}
\end{figure}

\noindent\textbf{Open-Field Deployment.}
This deployment uses ten sensing nodes in a larger outdoor area of approximately \(10{,}000~\mathrm{m}^2\). Compared with the interference-rich deployment, this site has fewer external acoustic interference sources although wind, ambient outdoor noise, and nearby buildings are still present. The terrain is mostly flat while the buildings create potential acoustic blockage and multipath effects. The target vehicle is an ATV-type vehicle following a nonuniform trajectory through the site. A sketch of this deployment is shown in Figure~\ref{fig:gq_map}.

\subsection{Results on the Interference-Rich Deployment}
\label{sub:res-int-rich}

We evaluate a set of algorithms with a fixed recommendation budget of \(|\mathcal S|=3\). We compare against the model-based selective-sensing framework of~\cite{buyukkalayci2025keeping}, which showed that low-cost acoustic Received Signal Strength Indicator (RSSI) measurements can be used to learn compact propagation models offline and use them for posterior-based sensor-subset recommendation online. From this line, we include the linear path-loss posterior and KDE-hybrid posterior baselines. The linear path-loss posterior uses a sensor-specific log-distance attenuation model, while the KDE-hybrid posterior uses empirical RSSI likelihoods when sufficient data is available and falls back to the path-loss model in data-sparse regions.

We also compare against the top-\(p\) sensor-selection framework of~\cite{buyukkalayci2026top}, which studies list-valued sensor recommendation as the primary objective and shows how normalized RSSI rules and geometry-aware posterior construction can support low-compute subset selection. From this line, we include the normalized RSSI top-3 baseline and the spline posterior baseline. The normalized RSSI baseline selects the three nodes with the largest normalized received signal strengths, while the spline posterior replaces the single global attenuation curve with a piecewise log-distance model to better capture distance-dependent RSSI behavior.

Finally, we include a Two-Tower (RSSI-only) ablation, which uses the same learned context--action architecture as the proposed method but replaces the frequency band powers with the overall RSSI for each node, reflecting the information used in the presented RSSI-based algorithms.

Table~\ref{tab:mean_accuracy} reports the mean closest-node containment accuracy over test timestamps on the interference-rich deployment, as defined in Section~\ref{sec:prob_for}. For the two-tower model, we use hidden width \(H=512\), two fully connected layers in each tower, and a one-hidden-layer prediction head. The utility weights are set to \((w_1,w_2,w_3)=(1,0.45,0.2)\). Empirically, the closest-node containment accuracy is not very sensitive to the choice of these weights, but assigning small nonzero weights to the selected sensors beyond the closest one gives the best results and makes training more stable. This sensitivity is reported in Table~\ref{tab:interference_weight_sensitivity_top1}, 
where the mean accuracy of the Two-Tower model with frequency-band features is given 
for different utility-weight configurations. Overall, the results show that the Two-Tower model 
with frequency-band features is substantially more robust to acoustic interference than 
the RSSI-only learned model and the model-based RSSI baselines.
\begin{table}[t]
\centering
\begin{threeparttable}
\caption{Mean closest-node containment accuracy in the interference-rich deployment.}
\label{tab:mean_accuracy}
\begin{tabular}{lcc}
\toprule
Method & Mean accuracy (\%) & Std. error (\%) \\
\midrule
\textbf{Two-Tower} & \textbf{98.39} & \textbf{0.33} \\
Two-Tower (RSSI-only) & 80.44 & 1.03 \\
Linear path-loss posterior & 77.03 & 1.09 \\
Normalized RSSI top-3 & 75.95 & 1.11 \\
KDE-hybrid posterior & 72.74 & 1.15 \\
Spline posterior & 71.33 & 1.17 \\
\bottomrule
\end{tabular}
\begin{tablenotes}[flushleft]
\footnotesize
\item All methods are evaluated on 1493 held-out test timestamps.
\end{tablenotes}
\end{threeparttable}
\end{table}

\begin{table}[t]
\centering
\footnotesize
\setlength{\tabcolsep}{2.5pt}
\renewcommand{\arraystretch}{0.95}
\begin{threeparttable}
\caption{Sensitivity of mean accuracy to utility weights in the interference-rich deployment.}
\label{tab:interference_weight_sensitivity_top1}
\begin{tabular}{@{}lccccc@{}}
\toprule
Weights & $(1,.45,.2)$ & $(1,1,1)$ & $(1,0,0)$ & $(1,.8,.6)$ & $(1,.8,.8)$ \\
\midrule
Mean acc. (\%) & \textbf{98.39} & 97.32 & 95.78 & 97.92 & 97.66 \\
\bottomrule
\end{tabular}
\begin{tablenotes}[flushleft]
\footnotesize
\item Accuracy is measured over 1493 held-out test timestamps using the Two-Tower model.
\end{tablenotes}
\end{threeparttable}
\end{table}

\begin{figure}[t]
    \centering
    \includegraphics[width=\columnwidth]{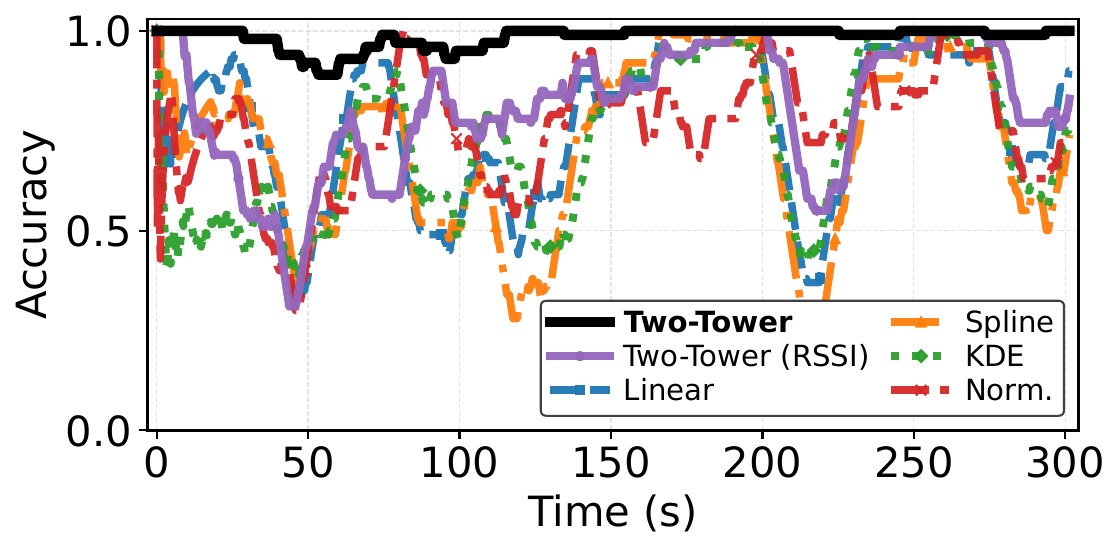}
    \caption{Rolling test accuracy over the test partition over a 100-sample window in the interference-rich deployment.}
    \label{fig:all_methods_time_accuracy}
\end{figure}

Table~\ref{tab:runtime_total} reports the corresponding online computation cost, including real-time feature construction and inference for each of the compared approaches. The frequency-band representation increases computation and storage relative to scalar RSSI methods, but the total runtime remains well below the \(200~\mathrm{ms}\) sensing interval. Thus, the additional spectral information improves containment accuracy while preserving real-time deployability. Additionally, Fig.~\ref{fig:all_methods_time_accuracy} shows closest-node containment accuracy over a rolling time window. While RSSI-based algorithms drop sharply in performance during intervals with interference, the Two-Tower model with integrated band-power features maintains substantially higher accuracy during those intervals.

\begin{table}[t]
\centering
\begin{threeparttable}
\footnotesize
\caption{Online computation time per timestamp in the interference-rich deployment.}
\label{tab:runtime_total}
\begin{tabular}{@{}lrrr@{}}
\toprule
Method & Feat. mean & Total mean & Total 99th pct. \\
       & (ms)       & (ms)       & (ms) \\
\midrule
Normalized RSSI top-3      & 0.0801 & 0.1074 & 0.3326 \\
Linear path-loss posterior & 0.0801 & 0.1366 & 0.4144 \\
Spline posterior           & 0.0801 & 0.1396 & 0.4351 \\
Two-Tower (RSSI-only)      & 0.0801 & 0.4817 & 1.2865 \\
\textbf{Two-Tower}         & \textbf{0.3560} & \textbf{0.7173} & \textbf{1.7026} \\
KDE-hybrid posterior       & 0.0801 & 0.8516 & 2.7207 \\
\bottomrule
\end{tabular}
\begin{tablenotes}[flushleft]
\footnotesize
\item Runtime is measured on a Windows 10 Pro 64-bit machine with an Intel Core i7-9750H CPU, 12 logical processors, and 16 GB RAM. Benchmarks are run on CPU using PyTorch over one thread. ``Feat. mean'' denotes the mean time for online feature construction, while ``Total mean'' includes feature construction, streaming state/context assembly where applicable, and decision computation. ``Total 99th pct.'' reports the 99th-percentile total computation time.
\end{tablenotes}
\end{threeparttable}
\end{table}

\subsection{Results on Open-Field Deployment}

Table~\ref{tab:graces_mean_accuracy} reports closest-node containment accuracy on the open-field deployment under the same evaluation protocol as Section~\ref{sub:res-int-rich}. On this site, every method performs substantially better than on the interference-rich deployment, with even the weakest baseline exceeding $92\%$, reflecting the cleaner acoustic conditions and the larger, more regular node layout. The RSSI-only Two-Tower model attains the highest accuracy ($99.40\%$), marginally ahead of the full frequency-band Two-Tower model ($97.80\%$), with both learned models outperforming the model-based posterior baselines. Table~\ref{tab:graces_runtime_total} reports the corresponding online computation times, which remain feasible within the $200~\mathrm{ms}$ sensing interval.

Comparing the two deployments reveals when the richer spectral representation can be worth the extra computational and storage cost. On the interference-rich deployment, the frequency-band Two-Tower model is dramatically more robust than its RSSI-only counterpart ($98.39\%$ vs.\ $80.44\%$): the band-power features let the model separate the target's acoustic signature from intermittent interference such as speech, wind, and passing vehicles, whereas using only RSSI is more likely to fail in distinguishing target-generated acoustic power from interference sources. The practical takeaway is that frequency-band features are most valuable precisely when the acoustic environment is contested; in benign conditions a lightweight RSSI-only model is sufficient and even slightly preferable.

\begin{table}[t]
\centering
\begin{threeparttable}
\caption{Mean closest-node containment accuracy in the open-field deployment.}
\label{tab:graces_mean_accuracy}
\begin{tabular}{lcc}
\toprule
Method & Mean accuracy (\%) & Std. error (\%) \\
\midrule
\textbf{Two-Tower (RSSI-only)} & \textbf{99.40} & \textbf{0.20} \\
Two-Tower & 97.80 & 0.38 \\
Spline posterior & 95.93 & 0.51 \\
Linear path-loss posterior & 93.19 & 0.65 \\
KDE-hybrid posterior & 92.99 & 0.66 \\
Normalized RSSI top-3 & 92.26 & 0.69 \\
\bottomrule
\end{tabular}
\begin{tablenotes}[flushleft]
\footnotesize
\item All methods are evaluated on 1498 held-out test timestamps.
\end{tablenotes}
\end{threeparttable}
\end{table}


\begin{table}[t]
\centering
\begin{threeparttable}
\footnotesize
\caption{Online computation time per timestamp in the open-field deployment.}
\label{tab:graces_runtime_total}
\begin{tabular}{@{}lrrr@{}}
\toprule
Method & Feat. mean & Total mean & Total 99th \\
       & (ms)       & (ms)       & pct. (ms) \\
\midrule
Normalized RSSI top-3      & 0.0011 & 0.0095 & 0.0376 \\
Linear path-loss posterior & 0.0011 & 0.1036 & 0.2909 \\
Spline posterior           & 0.0011 & 0.1358 & 0.3136 \\
\textbf{Two-Tower (RSSI-only)} & \textbf{0.0011} & \textbf{0.4453} & \textbf{1.3542} \\
KDE-hybrid posterior       & 0.0011 & 0.9520 & 3.0650 \\
Two-Tower                  & 0.6845 & 1.0854 & 2.5439 \\
\bottomrule
\end{tabular}
\begin{tablenotes}[flushleft]
\footnotesize
\item Runtime is measured on a Windows 10 Pro 64-bit machine with an Intel Core i7-9750H CPU, 12 logical processors, and 16 GB RAM. Benchmarks are run on CPU using PyTorch over one thread.
\end{tablenotes}
\end{threeparttable}
\end{table}

\section{Computational Complexity}

Posterior-based scalar-RSSI methods compute a likelihood for every spatial hypothesis. For one target, the dominant online cost is \(O(|\mathcal H|V_\ell)\), where \(|\mathcal H|\) is the number of grid locations and \(V_\ell\) is the number of low-cost acoustic nodes. For \(M\) jointly modeled targets, direct posterior inference becomes \(O(|\mathcal H|^M V_\ell)\). The cost is therefore dominated by the latent spatial grid, and multi-target inference quickly becomes expensive if no additional assumptions narrowing the search space are present.

For fixed audio window length, band count, and network width, the online cost of the Two-Tower method is
\[
O(V_\ell + V_h^K),
\]
where \(V_\ell\) is the number of low-cost acoustic nodes, \(V_h\) is the number of recommendable high-cost assets, and \(K\) is the subset-size budget. The \(O(V_\ell)\) term results from constructing and transmitting the network-wide acoustic context, while the \(O(V_h^K)\) term results from exact scoring of candidate subsets. Thus, the Two-Tower model removes dependence on the spatial hypothesis grid and joint multi-target posterior enumeration.

On the other hand, a naive scaling of the Two-Tower inference becomes 
costly as \(V_h\) increases. In larger deployments, this can be addressed by replacing exhaustive enumeration with utility aggregation over smaller subsets or individual nodes, hierarchical narrowing from larger to smaller regions, two-stage retrieval/reranking, or graph-based subset encoders.

\section{Conclusion}

In this paper, we formulate selective sensor activation as a recommendation problem,
where network-wide acoustic observations define the context and candidate sensor subsets
define the items to be ranked. Using a Two-Tower architecture and frequency-band acoustic
features, we show that the proposed method improves robustness to acoustic interference.
Through experiments, we also show that the model outperforms previous approaches while remaining
computationally feasible for real-time tracking.
More broadly, our results suggest that recommendation-style architectures can be adapted to
efficient sensor-subset selection.

\balance
\bibliographystyle{IEEEtran}
\bibliography{reference}

\end{document}